# Enhancing Small Object Detection with YOLO: A Novel Framework for Improved Accuracy and Efficiency

Mahila Moghadami[1], Mohammad Ali Keyvanrad[1*] and Melika Sabaghian[1]

[1] Faculty of Electrical & Computer Engineering Malek Ashtar University of Technology, Iran
*keyvanrad@mut.ac.ir

**Abstract.** This paper investigates and develops methods for detecting small objects in large-scale aerial images. Current approaches for detecting small objects in aerial images often involve image cropping and modifications to detector network architectures. Techniques such as sliding window cropping and architectural enhancements, including higher-resolution feature maps and attention mechanisms, are commonly employed. Given the growing importance of aerial imagery in various critical and industrial applications, the need for robust frameworks for small object detection becomes imperative. To address this need, we adopted the base SW-YOLO approach to enhance speed and accuracy in small object detection by refining cropping dimensions and overlap in sliding window usage and subsequently enhanced it through architectural modifications. we propose a novel model by modifying the base model architecture, including advanced feature extraction modules in the neck for feature map enhancement, integrating CBAM in the backbone to preserve spatial and channel information, and introducing a new head to boost small object detection accuracy. Finally, we compared our method with SAHI, one of the most powerful frameworks for processing large-scale images, and CZDet, which is also based on image cropping, achieving significant improvements in accuracy. The proposed model achieves significant accuracy gains on the VisDrone2019 dataset, outperforming baseline YOLOv5L detection by a substantial margin. Specifically, the final proposed model elevates the mAP .5.5 accuracy on the VisDrone2019 dataset from the base accuracy of 35.5 achieved by the YOLOv5L detector to 61.2. Notably, the accuracy of CZDet, which is another classic method applied to this dataset, is 58.36. This research demonstrates a significant improvement, achieving an increase in accuracy from 35.5 to 61.2.

## 1 Introduction

Object detection plays a critical role in artificial intelligence research, particularly within the field of machine vision. In this context, the primary goal of object detection is to identify and draw a bounding box — the optimal rectangle that encompasses all parts of an object while minimizing the inclusion of irrelevant background elements. This task must be accomplished across all objects present in a given scene. Over the years, the field has made significant advancements, driven by the rapid growth of data availability, improved computational power, and the development of more sophisticated AI algorithms.

Object detection has a wide range of applications across diverse sectors, including industrial automation, medical imaging, military surveillance, and security systems. Notable uses include identity recognition, assisting the visually impaired with object identification, and enabling autonomous vehicles. Despite the impressive results achieved by recent algorithms, such as the latest versions of YOLO and transformer-based detectors, their performance in aerial imagery remains suboptimal, both in terms of efficiency and accuracy.

Aerial imagery presents several unique challenges, primarily due to the non-uniform distribution and small size of objects within large images. These challenges affect both the training and inference stages of object detection algorithms. Specifically, detecting small objects in large aerial images is complicated by the image resizing process, which is necessary for input into the object detection network. Resizing often leads to a significant reduction in image size, which diminishes the apparent size of small objects. As a result, the effectiveness of the feature map is reduced, negatively impacting detection accuracy.

Moreover, aerial images present additional complexities such as densely packed small objects with high overlap, diverse scales of objects within the image, and unbalanced samples of objects. These intricacies further exacerbate the challenges faced in effectively detecting objects within aerial imagery.

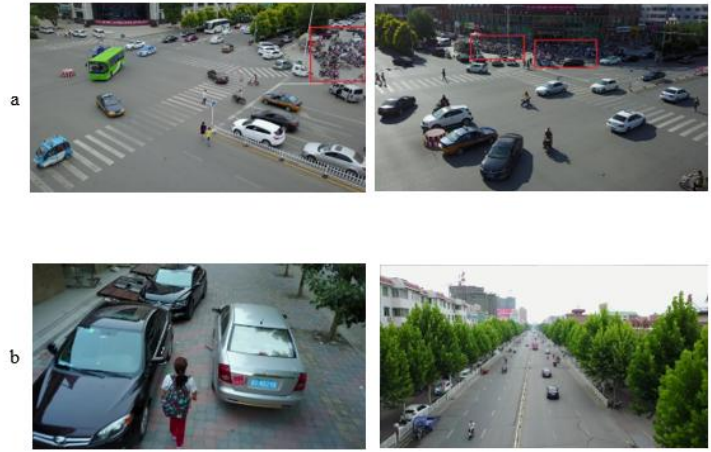

**Fig. 1.** Examples of challenges in aerial imagery: (a) dense small objects with significant overlap, and (b) the wide range of object scales within the dataset.

According to the more conventional definition provided [1] objects with dimensions smaller than $32 \times 32$ pixels are classified as small objects, regardless of the input image size.

Based on the considerations outlined above, summary of the activities conducted in this research has been categorized as follows:

- Evaluation of SAHI[2] parameters and optimization for performance enhancement on the datasets under investigation. After analyzing the results from each parameter, the optimal settings for the datasets used in this study were determined.
- The CZDet[3] method was evaluated as the baseline approach, and the effect of applying super-resolution technology during both training and inference was examined.
- Finally, the SW-YOLO[4] base model was analyzed, and improvements were made to its head, neck, and backbone components. These enhancements resulted in increased accuracy for both the base and final models proposed in this research.

The paper begins with a comperhensive review of prior works, summarizing existing approaches to the challenges identified in this research,small object detection and the analysis of large-scale images. Subsequently, the discussion moves to SAHI parameter optimization, where an in-depth examination is undertaken. Efforts to improve the CZDet framework are also analyzed. A detailed, step-by-step account of enhancements made to SW_YOLO is provided. Finally, the accuracy and speed of the examined methods are reported and compared, showcasing the innovations applied to the VisDrone dataset.

## 2    Related works

Based on the investigations conducted, the solutions to the two primary challenges of this research—small object size and large image scale—can be categorized as follows. While this study initially examines the solutions to each challenge separately, the final framework and method proposed are inspired by integrating solutions to both challenges.

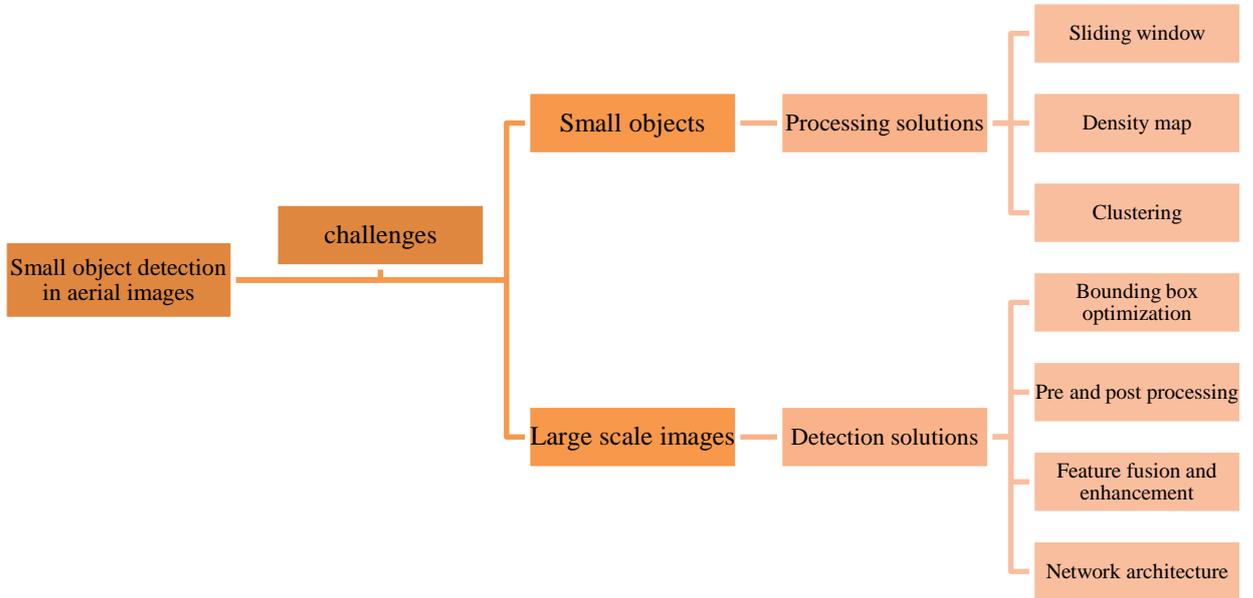

**Diagram 1**. A review of the existing general approaches in the field of small object detection in aerial images.

## 2.1 Small object detection

Solutions for detecting small objects can be grouped into three main categories: network architecture, pre-processing and post-processing methods, and bounding box optimization techniques. This section reviews the studies conducted in each of these areas to assess their effectiveness in addressing the challenge of small object detection.

**Network Architecture.** Multi-scale object detection methods are categorized into four main classes: image pyramid, prediction pyramid, integrated feature, and feature pyramid[5].

- **Image Pyramid** uses images at different scales as input to the network for object detection and recognition.
- **Prediction Pyramid** involves utilizing feature maps at different scales for prediction.
- **Integrated Feature** makes predictions based on a feature map obtained by combining feature maps from multiple scales.

- **Feature Pyramid** merges the prediction pyramid and integrated feature approaches, extracting information at multiple dimensions and sizes for prediction tasks.

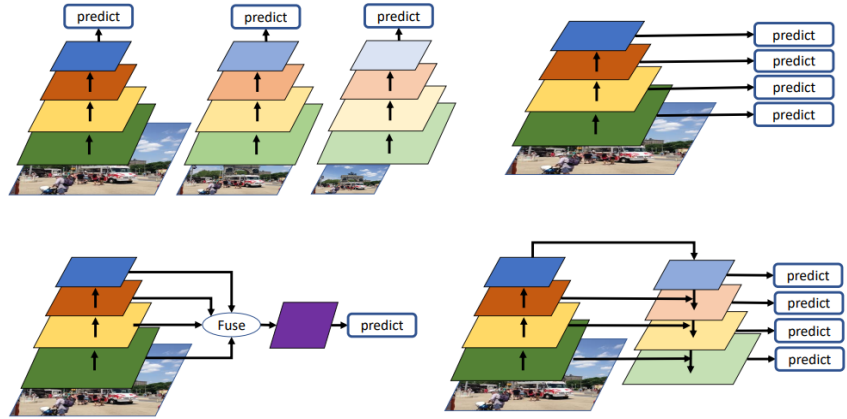

**Fig. 2.** Four paradigms for multi-scale feature learning. Top Left: Image Pyramid; Top Right: Prediction Pyramid; Bottom Left: Integrated Features; Bottom Right: Feature Pyramid.

One of the earliest improvements in multi-scale detection was introduced by YOLOv3[6], which incorporated a prediction pyramid to enhance detection across different scales. TridentNet[7] further advanced this concept by adopting a multi-branch detection approach, combining image and feature pyramids. Instead of relying on multiple input images, TridentNet utilized parallel branches to generate feature maps at varying scales, improving detection performance.

Several studies have explored modifications to network architectures to enhance small object detection. In [8], a novel detection head was introduced to generate higher-resolution feature maps, along with an attention mechanism integrated at the end of the backbone. This approach reduced computational overhead while preserving crucial spatial information. Furthermore, a new loss function was proposed to enhance detection accuracy. Similarly, in [9], modifications to the YOLO architecture were explored by replacing the conventional four detection heads with transformer-based heads (Transformer Prediction Head, TPH). This self-attention mechanism refined prediction accuracy, while the Convolutional Block Attention Module (CBAM) was employed to emphasize critical regions in densely populated scenes. The study also leveraged data augmentation, multi-scale evaluation, and model ensembling to further improve detection performance. The TPH-YOLOv5 architecture, illustrated in the figure below, enhances tiny object detection for the VisDrone2021 dataset

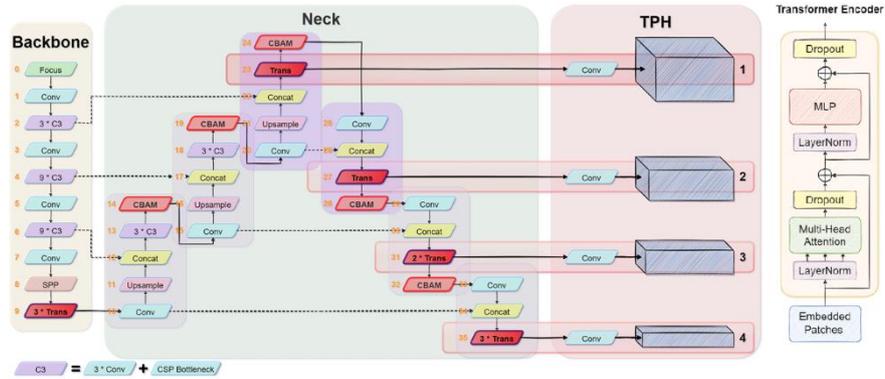

**Fig. 3.** The architecture of the TPH-YOLOv5 [10].

Despite the advantages of additional detection heads, their computational and memory overhead remains a challenge. To address this, TPH-YOLOv5++ [11] introduced the CATrans module, an alternative to multiple detection heads, which preserved high-level feature information while maintaining computational efficiency. Similarly, in the HIC-YOLO framework [8], a redesigned detection head, combined with the CBAM module in the backbone, was proposed to enhance detection accuracy for small objects. Another study [12] introduced structural modifications to YOLOv5s by replacing PANet[13] with BiFPN[13] in the neck and proposing a new loss function tailored for small object detection. Moreover, in [14], detection accuracy was improved by integrating a bottleneck block into the backbone, enabling better feature extraction from shallow layers. This study also introduced a redesigned detection head along with other architectural optimizations to further enhance small object detection.

These approaches collectively highlight the ongoing advancements in object detection, particularly in improving small object recognition through architectural modifications, attention mechanisms, and optimization strategies.

**Feature Fusion and Enhancement.** One of the key advancements in improving feature extraction for small object detection is the use of feature fusion networks, such as the FPN[15]. FPNs improve the quality of feature maps and are commonly used as the neck component in many detection architectures. Enhanced versions of FPN, such as PAFPN[13], NasFPN[16], and ImFPN[17], focus on further improving feature fusion. In [18], a novel HR (High-Resolution) block was introduced for effective feature fusion. In this block, convolutional operations with different kernel sizes are applied at each layer, generating feature maps that combine strong semantic information and fine details at different scales. These feature maps are then fused to enhance small object detection. Additionally,[19] proposed a method specifically for small object detection in aerial imagery, aiming to increase semantic information from shallow layers of feature maps.

Gold-YOLO[20], utilizing the Gather-and-Distribute mechanism, enhances accuracy by fusing multi-scale feature maps. This mechanism gathers global information from all primary levels, combines it, and returns it to each level for improved detection. Other studies, such as PPYOLO[21] and PPYOLOE[22], have focused on improving accuracy by modifying the neck component and refining feature map fusion strategies.

The following figure illustrates the differences between various feature extraction methods utilized for small object detection. This image provides a clear visual representation of the proposed techniques, including Feature Fusion Networks like FPN, PAFPN, and others, highlighting their unique approaches and effectiveness in enhancing detection accuracy.

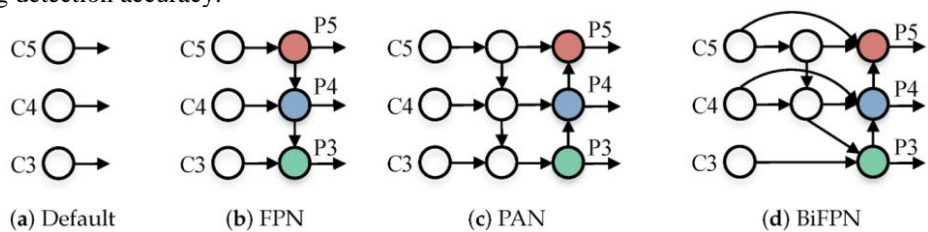

**Fig. 4.** common feature fusion paths [11].

## 2.2 Large-scale images

Several methods for processing large-scale images and improving small object detection have been proposed. These include the sliding window approach, density map method, and clustering approach.

**Sliding Window.** The sliding window method involves dividing the image into overlapping slices, which are then processed by the object detection network. The results from all slices are combined to produce a final detection output. While this method improves accuracy, it significantly increases computational time, making it less suitable for real-time applications. The SAHI [23] framework is one of the robust and effective works in this field, primarily focusing on the sliding window technique, which can be utilized during both the training and inference stages. Moreover, study [47] proposed a framework that optimizes computational cost and reduces inference time compared to existing approaches. In this research, the slice dimensions are correlated with the input image size, allowing the model parameters for the sliced data to be maintained in consistent ratios with the main dataset. Several tiling-based methods have been developed to enhance object detection, particularly for small objects in challenging scenarios. For instance, the EdgeDuet framework [24] utilizes a series of key steps, including tile-level parallelism, to process video frames more efficiently by decompressing blocks containing no small objects and optimizing detection through overlap-tiling. Additionally, another study [25] focuses on detecting pedestrians and vehicles from micro aerial vehicles using high-resolution imagery, employing a tiling approach that improves accuracy during both training and inference phases. This method effectively reduces detail loss while ensuring the model receives fixed-size inputs, showcasing significant improvements in performance on platforms like Nvidia Jetson TX1 and TX2 with the VisDrone2018 dataset. These techniques reflect the ongoing innovations in object detection, aiming to overcome the limitations of conventional processing methods.

**Density Map.** This method generates a density map to identify regions of high object concentration in the image. Slices are determined based on these density regions, and object detection is performed on each slice. Compared to the sliding window method, the density map approach reduces computational costs while still enabling effective object detection. The Object-Activation Network introduced in [26] uses image slices to output object activation maps, processing only those slices with object densities above a certain threshold to optimize computational efficiency.

**Clustering.** Another method for detecting small objects in large-scale images is clustering. A study by [3] presented a clustering-based approach to identify dense regions in images, referred to as "dense area slices." These regions are processed separately to improve small object detection accuracy. Additionally, modules like ScaleNet and PP ensure consistency across object scales. The GLSAN[27] framework was developed to enhance small object detection in densely populated regions. It includes three main modules: GLDN for general and localized object detection, SARSA for clustering dense areas using K-means, and LSRN, which improves the quality of regions identified by SARSA before passing them to the detection network.

A notable clustering-based approach, the Clustered Detection (ClusDet) network introduced in [28], addresses the challenges of detecting small, sparsely, and non-uniformly distributed objects in aerial images. ClusDet unifies object clustering and detection into an end-to-end framework. It comprises a cluster proposal sub-network (CPNet) that identifies object cluster regions, a scale estimation sub-network (ScaleNet) that estimates object scales for these regions, and a dedicated detection network (DetecNet). This method significantly reduces the number of image chips required for final object detection by focusing only on predicted cluster regions, thereby optimizing computational efficiency. Furthermore, the cluster-based scale estimation in ClusDet improves the accuracy of small object detection compared to single-object based methods, and the DetecNet leverages contextual information within these clustered regions to boost overall detection accuracy.

Further refining clustered object detection, a study in [29] proposes an improved cluster chip selection scheme. This method enhances detection performance in aerial images by more effectively identifying "cluster chips"—dense object regions—and applying fine-grained detectors to them

In the next chapter, the general concept of this study is presented along with a comprehensive explanation of the classical implementation method used for comparison. Additionally, the optimization of SAHI parameters is discussed to evaluate their performance against the final results.

# 3    Methodology

## 3.1    SAHI parameter optimization

The Slicing Aided Hyper Inference (SAHI) framework is designed to improve small object detection in large-scale images through two main pipelines: **model training** and **inference**.

- **Model Training:** During the training process, images are divided into slices with specific dimensions and overlaps to make better use of pre-trained models. This approach effectively increases the number of training images, enhancing the model's accuracy.

- **Inference:** In the inference phase, the original image is also segmented into slices and, along with the complete original image, is passed through the trained network. To remove redundant predictions, several merging methods, including Non-Maximum Suppression (NMS), Local Soft-NMS (LSNMS), Non-Maximum Merging (NMM), and Greedy NMM, are applied. These methods compare bounding boxes based on confidence scores and overlap, helping to ensure accurate object detection.

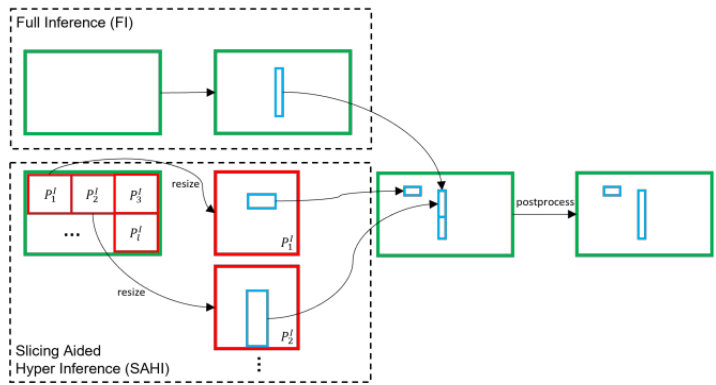

**Fig. 5.** Inference with slicing window[2].

Additionally, **Intersection over Union (IoU)** and **Intersection over Self (IoS)** metrics are utilized during post-processing to improve adaptability in the testing phase.

## 3.2    CZDet improvement

Existing methods for object detection in cluttered areas often rely on segmenting densely populated regions or clustering techniques, which can be time-consuming due to the need for additional trainable modules or processing units. To address this, CZDet[3] proposes a solution where the detection network itself identifies cluttered regions, avoiding extra modules. These identified regions are then re-evaluated with higher precision, enhancing detection accuracy for smaller objects. Training and inference pipeline is shown completely in **Fig. 6.**

Analysis revealed that images labeled as "cut" class outputs, whether included in the dataset or reprocessed during inference, often have significantly reduced dimensions compared to the originals. Detection networks are typically trained on images averaging $800 \times 800$ pixels, whereas "cut" images may be around $200 \times 250$ pixels. Consequently, these images need resizing to meet the network's input requirements. Traditional resizing, often done through interpolation, can degrade image quality, resulting in blurriness and loss of critical details.

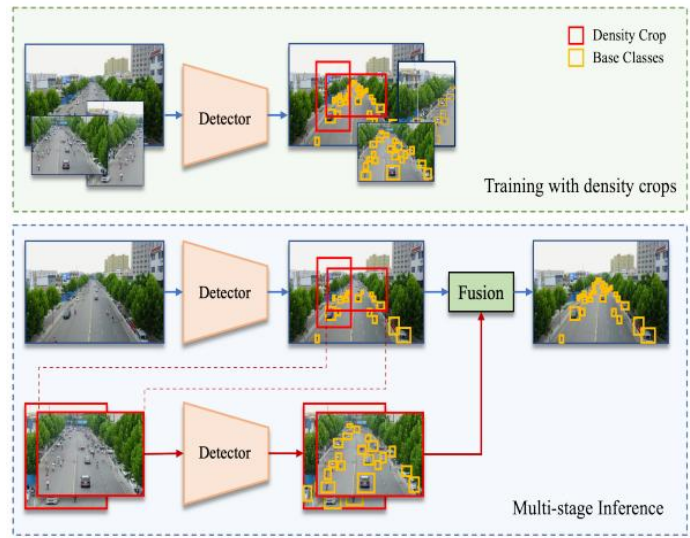

**Fig. 6.** Training and inference pipeline in CZDet [3].

To counteract this, an initial solution involved incorporating a super-resolution network architecture to enhance image quality. SR networks are designed to generate high-resolution images from low-resolution counterparts. An SR module was integrated into the network architecture to support both training and testing:

- **Training Phase:** "Cut" images are processed through the SR module to produce higher-quality versions, which are then used to train the network. This enhancement, however, significantly extends training duration.

- **Testing Phase:** During testing, images predicted as "cut" class by the detection network undergo SR processing before re-entering the detection network. To optimize this process, an SR model specifically trained on the target dataset was used, transforming low-resolution images into high-resolution versions. This approach increases detection accuracy for smaller objects by enhancing image resolution during both training and testing.

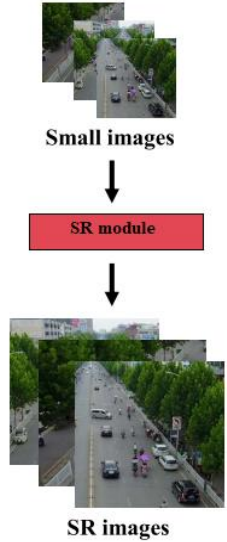

**Fig. 7.** Super resolution module.

### 3.3 SW-YOLO enhancement

SW_YOLO[4], proposed an efficient drone object detection framework that addresses challenges like dense clusters, overlapping objects, and scale diversity. Their approach uses a uniform slicing window method, dividing input images into smaller patches for detecting small objects while maintaining efficiency. The framework includes global detection on the full image and local detection on sub-patches to handle objects at different scales. A scale filtering mechanism assigns objects to the appropriate detection task to maintain scale invariance. Additionally, the method uses random anchor-cropping for data augmentation, enriching training data with diverse scenarios.

Two customized augmentations simulate real-world scenarios with dense object clusters, particularly aiding the detection of rare categories. Comprehensive experiments demonstrated that this approach significantly improves detection performance with lower computational costs compared to other methods. SW_YOLO workflow is shown in  This framework serves as a baseline for evaluating our proposed approach.

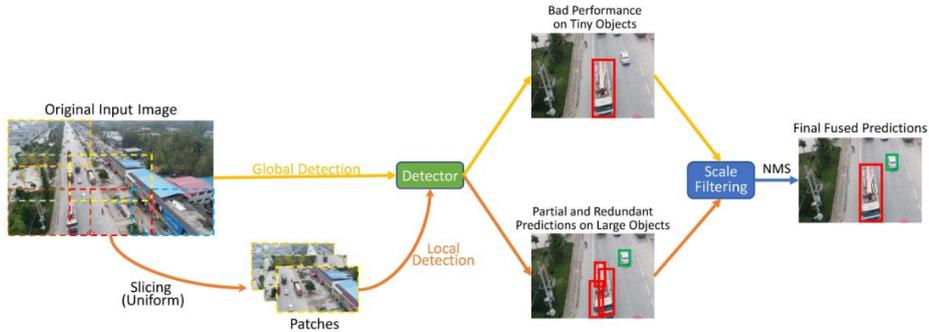

**Fig. 8.** SW_YOLO workflow[4].

This study introduces several enhancements to the YOLO architecture, specifically optimized for small object detection. Inspired by [8], these improvements, designed to boost both accuracy and computational efficiency, are implemented at three main levels: adding a new detection head for small objects, integrating the Convolutional Block Attention Module (CBAM), and utilizing an involution block for advanced feature extraction. Base yoloV5 architecture is shown in **Fig. 9.** These architectural modifications are integrated into the YOLOv5 base model as follows, with the final enhanced YOLOv5 architecture presented in **Fig. 14**.

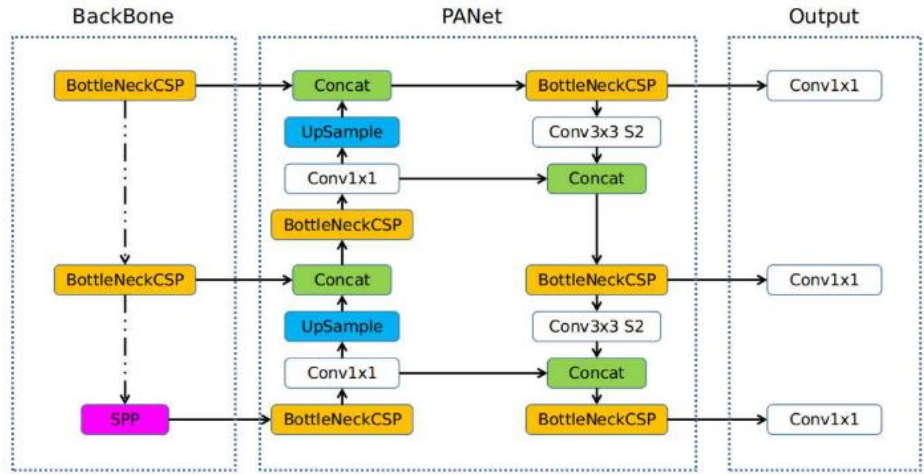

**Fig. 9.** Overview of yoloV5 architecture [30].

1. **Adding a New Detection Head for Small Objects**
   Research shows that increasing the resolution of feature maps enhances small object detection accuracy. To exploit this, beyond the typical feature maps (P3, P4, and P5) used in YOLO for detecting small, medium, and large objects respectively, this framework includes an additional high-resolution

P2 layer (160×160 pixels). The added P2 layer captures more detailed features, making it ideal for detecting small objects with precision.

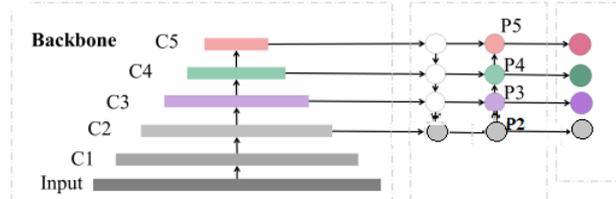

**Fig. 10.** The structure of the YOLO network after adding the new head.

## 2. CBAM Attention Mechanism

To prioritize critical spatial and channel-wise information, the CBAM module is incorporated at the tail end of the backbone. While conventional approaches often place CBAM in the network's neck section, positioning it within the backbone minimizes computational overhead due to smaller feature map dimensions (20×20) at this stage. CBAM consists of two attention blocks, CAM (Channel Attention Module) and SAM (Spatial Attention Module), each targeting different aspects of feature optimization.

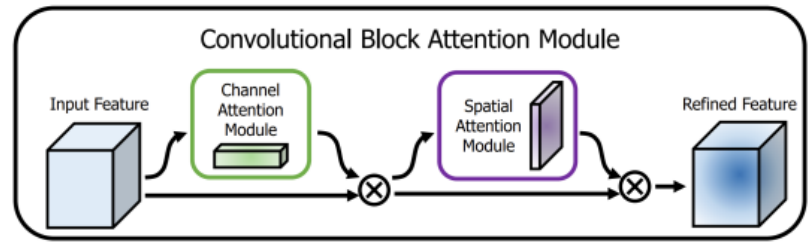

**Fig. 11.** CBAM architecture[31].

- o **CAM**: This module captures channel-specific importance by aggregating spatial information using both average and max pooling, then generating a channel attention map through a lightweight neural network. This attention map applies unique weights to each channel, optimizing the relevance of channel-specific features.

- o **SAM**: Following CAM, SAM emphasizes critical spatial locations. It utilizes pooling operations to reduce dimensions and a 7×7 convolutional layer to create a spatial attention map, assigning greater weights to essential regions in the image.

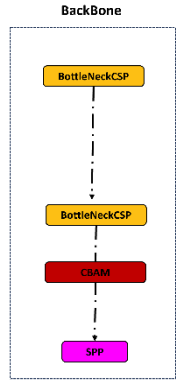

**Fig. 12.** Adding new block CBAM in backbone.

The newly added block has been incorporated into the image and highlighted in red.

3. **Involution Block**

The involution block replaces traditional convolution layers to refine spatially relevant feature extraction. Unlike fixed convolution filters (spatial-agnostic), involution uses dynamic, spatially specific filters, applying tailored filters to each location in the image. This allows the network to better retain location-specific information.

Within the involution block, a unique kernel is generated per pixel, applied uniformly across all channels. This kernel is then combined with the input feature maps through convolution. Finally, a summation aggregation step consolidates the extracted features across neighboring pixels, preserving spatial context and enhancing detection precision.

**PANet**

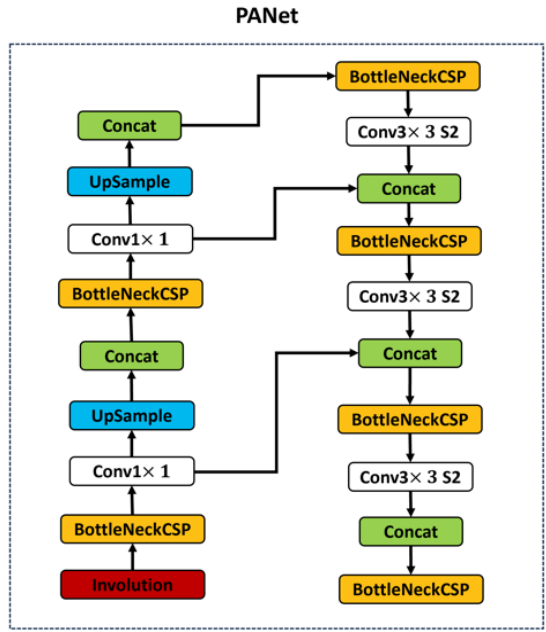

**Fig. 13.** Adding new block involution in backbone.

Collectively, these strategies significantly boost YOLOv5's capability for small object detection. By reducing computational load and increasing detection accuracy, these enhancements make YOLOv5 more suitable for industrial applications demanding both high speed and reliability.

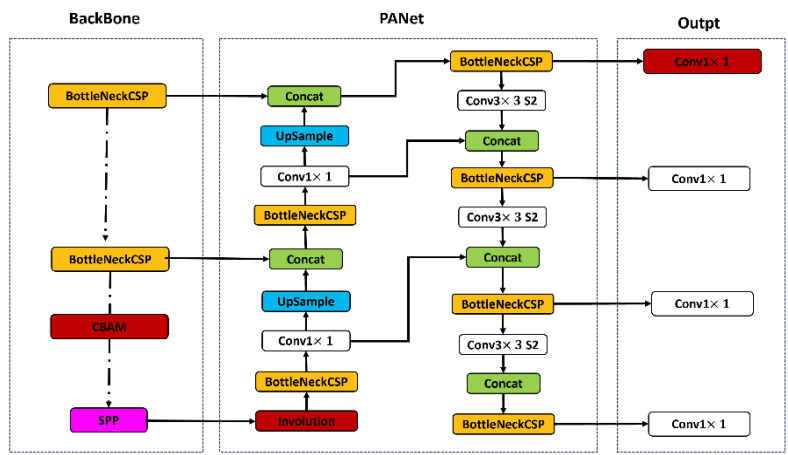

**Fig. 14.** The final architecture presented.

The newly added block has been incorporated into the image and highlighted in red.

In the next chapter, the utilized dataset is introduced, followed by the presentation of the results for each proposed idea and a comparison with previous methods.

# 4    Experimental results

## 4.1    Dataset

In this study, the VisDrone-Det2019 dataset has been selected for training and evaluation in object detection tasks. As a subset of the broader VisDrone challenge, this dataset specifically addresses object detection in static images and comprises 6,471 training images with resolutions ranging from 1920×1080 to 3840×2160, representing 10 distinct object classes. Notably, approximately 31.25% of objects in this dataset are categorized as "small" ($> 32^2$), highlighting the challenge of detecting smaller objects within high-resolution images.

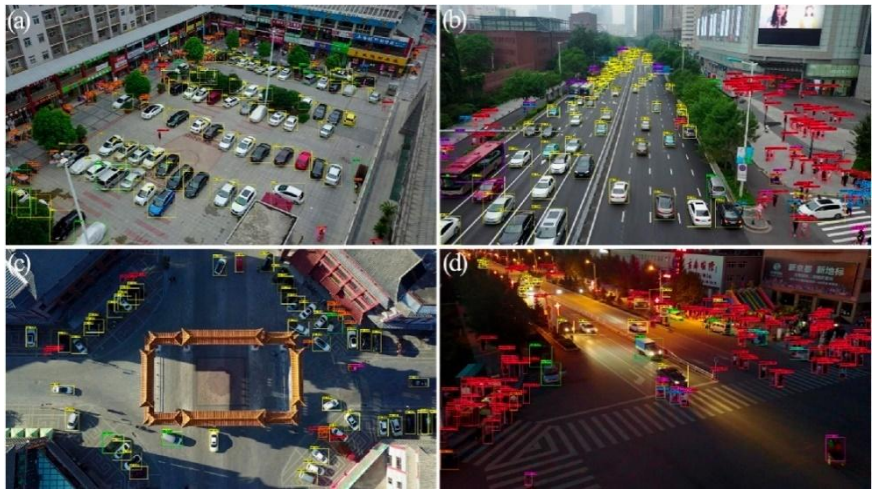

**Fig. 15.** Examples of images from the VisDrone2019 dataset.

## 4.2    Result

This section elucidates the findings derived from the training of baseline object detection models, which were undertaken to establish a robust foundation for subsequent experimentation. The models were initially pre-trained on the MS-COCO dataset over the course of 300 epochs, employing a batch size of 32. Evaluation of model performance was conducted utilizing COCO metrics, specifically focusing on Average Pre-cision (AP) at various Intersection over Union (IoU) thresholds, thereby ensuring the consistency of accuracy measures. The YOLO-based models, distinguished by their single-stage architecture, demonstrated markedly superior inference speeds in comparison to two-stage models, such as Faster R-CNN, which necessitates an additional stage for region proposal generation. Among the various YOLO architectures, YOLOv5L was identified as the primary baseline model owing to its robust performance, achieving an AP of 47.3% at an IoU threshold of 0.5, thereby illustrating a commendable equilibrium between accuracy and processing speed. To further enhance the accuracy of small-object detection, the SAHI (Slicing Aided Hyper Inference) approach was implemented. This method focuses on segmenting images to improve detection performance, particularly for small objects that often evade detection in full-frame evaluations. Various post-processing techniques were assessed with SAHI, including adjustments to overlap ratios and crop sizes. The optimal strategy combined full-image predictions with those derived from image slices, yielding a significant enhancement in the model's capability to detect small objects, while maintaining an overall precision of 65.1% across all object sizes. However, this approach introduced a trade-off, as the complexity of processing both full and cropped images resulted in a measurable reduction in inference speed, decreasing from 30 FPS to 18 FPS.

Further advancements were explored through two primary strategies targeting the optimization of baseline models. The first strategy involved the pretraining of the CZDet model using ImageNet weights, which led to a considerable increase in accuracy, achieving an AP of 50.5% on the validation set. However, the introduction of a Super-Resolution (SR) module during training unexpectedly resulted in a decline in accuracy, with the AP dropping to 45.2% and an increase in training time. This decline is hypothesized to stem from noise amplification in images that were either blurry or of low resolution, particularly in challenging night scenes. The second strategy centered on refining the SW-YOLO model by integrating additional modules, specifically the Convolutional Block Attention Module (CBAM) and Involution. These modules were selected for their ability to enhance feature representation without imposing excessive computational demands. CBAM's attention mechanisms facilitated improved focus on critical regions of interest, enhancing the model's ability to detect small and obscured objects across various scales.

The incorporation of CBAM and Involution into the SW-YOLO architecture resulted in significant improvements in both accuracy and robustness. The optimized SW-YOLO model achieved an AP of 52.7% at IoU=0.5, surpassing the performance of standard single-stage detectors while maintaining a relatively stable inference speed of approximately 25 FPS. This trade-off between accuracy and speed remains favorable for a wide range of applications, as the SW-YOLO model effectively balances computational efficiency with enhanced detection precision. The strategic integration of CBAM and Involution empowered SW-YOLO to leverage detailed contextual information, ultimately making it a highly effective choice for scenarios requiring both rapid processing and high-accuracy object detection.

**Table 1.** Evauation results on VisDrone dataset.

| Model | AP 0.5 | APsmall | FPS |
|---|---|---|---|
| FasterRCNN | 41.3 | - | - |
| YOLOv5S | 31.4 | 22.3 | 29.66 |
| YOLOv5M | 35.6 | 10.5 | 32.2 |
| YOLOv5L | 35.5 | 11.1 | 31.3 |
| YOLOv5s-FineTune | 39 | 15.3 | 29.66 |
| YOLOv5L-FineTune-SAHI | 48.3 | 36.7 | 7.14 |
| CZDet | 58.36 | 25.86 | 4.64 |
| CZDet-SR-Train | 56.34 | 24.43 | - |
| CZDet-SR-Inference | 58.13 | 25.71 | - |
| SWYOLO | 60.4 | 29.2 | 21.92 |
| SWYOLO-CBAM-Involution-p2 | **61.2** | **30** | 18.42 |

The baseline SW-YOLO model initially achieved an accuracy of 60.4% mAP0.5. Various modifications were subsequently applied to this foundational architecture to optimize performance. Introducing a transformer module—by adding the C3TR module into the backbone and individually into each detection head—resulted in reductions in both accuracy and processing speed. Additionally, the original VIOU loss function was replaced with the SIOU loss function used in this implementation; however, this substitution did not yield accuracy improvements.

Further analysis examined the effects of incorporating a new head, the Involution block, and the CBAM block as standalone modifications. While the CBAM and Involution blocks did not enhance accuracy, they contributed to higher processing speeds, increasing the FPS of the baseline model by 3.78 and 2.68, respectively. Ultimately, the integration of these modules, combined with the new detection head, resulted in accuracy gains while maintaining a competitive processing speed, with only a 3.5-unit reduction relative to the original SW-YOLO model. This enhanced configuration outperformed the fastest baseline detectors, achieving a 0.57-unit increase in speed and 1.7 times higher accuracy.

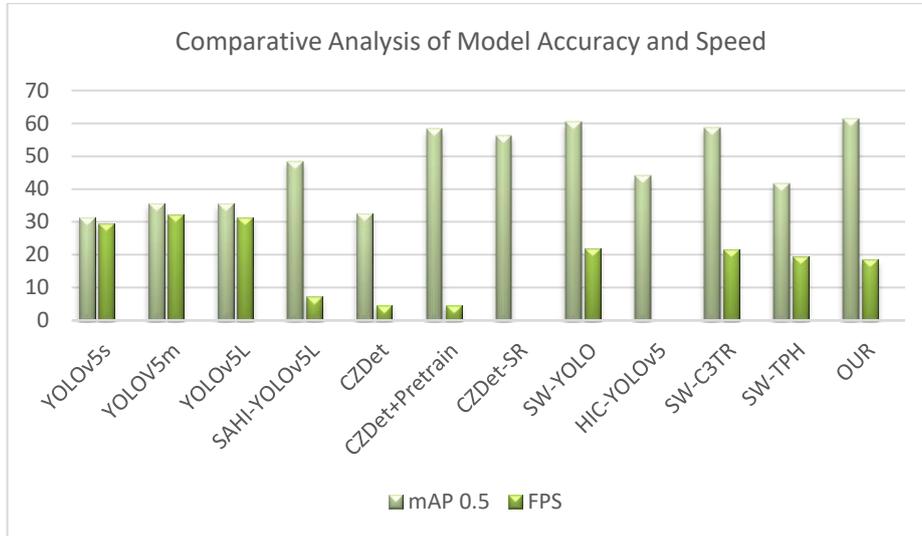

**Fig. 16.** Comparison Chart of Accuracy and Speed of Various Models.

## Conclusion

This thesis proposes a framework for small-object detection in large-scale images, aiming to balance inference speed and accuracy effectively. The approach employs an image slicing technique to improve the detection of small objects by generating high-resolution image segments. This technique is evaluated both in the training phase, to expand the dataset, and during inference, to enhance detection accuracy. Various post-processing strategies were tested to integrate these image slices, with the IOS and NMS methods yielding the most favorable results. Furthermore, incorporating the full image alongside the segmented slices significantly improved precision, particularly for larger objects.

For model selection, a high-performing baseline was chosen based on recent advancements evaluated on the VisDrone2019 dataset. To enhance small-object detection, a Super-Resolution network was incorporated at both the inference and training stages, increasing the clarity of images containing densely packed small objects.

Additional refinements included integrating the CBAM into the backbone network to focus on critical spatial and channel features with minimal computational overhead. The use of Involution blocks in the neck module further strengthened feature map quality, while an additional detection head was added to leverage higher-resolution feature maps, ultimately improving small-object detection performance.